\documentclass[10pt,twocolumn,letterpaper]{article}

\usepackage{cvpr}      %
\usepackage[T1]{fontenc}

\definecolor{cvprblue}{rgb}{0.21,0.49,0.74}
\usepackage[pagebackref,breaklinks,colorlinks,allcolors=cvprblue]{hyperref}
\def\our{VeGaS}
\def\ourdens{Folded-Gaussians}

\def\G{\mathcal{G}}

\def\N{\mathcal{N}}

\def\m{\mathrm{m}}
\def\v{\mathrm{v}}
\def\x{\mathrm{x}}

\def\e{\mathrm{e}}
\def\r{\mathrm{r}}
\def\s{\mathrm{s}}

\def\R{\mathbb{R}}

\def\ga{\mathcal{T}}

\def\paperID{6612} %
\def\confName{CVPR}
\def\confYear{2025}

\title{\our{}:  Video Gaussian Splatting}

\author{
Weronika Smolak-Dyżewska$^*$,  Dawid Malarz$^*$, Kornel Howil$^*$, \\
Jan Kaczmarczyk, Marcin Mazur, Przemysław Spurek\\
Jagiellonian University\\
Faculty of Mathematics and Computer Science\\
{\tt\small weronika.smolak@doctoral.uj.edu.pl}
}

\begin{document}

\twocolumn[{%
\maketitle
\includegraphics[width=.99\linewidth]{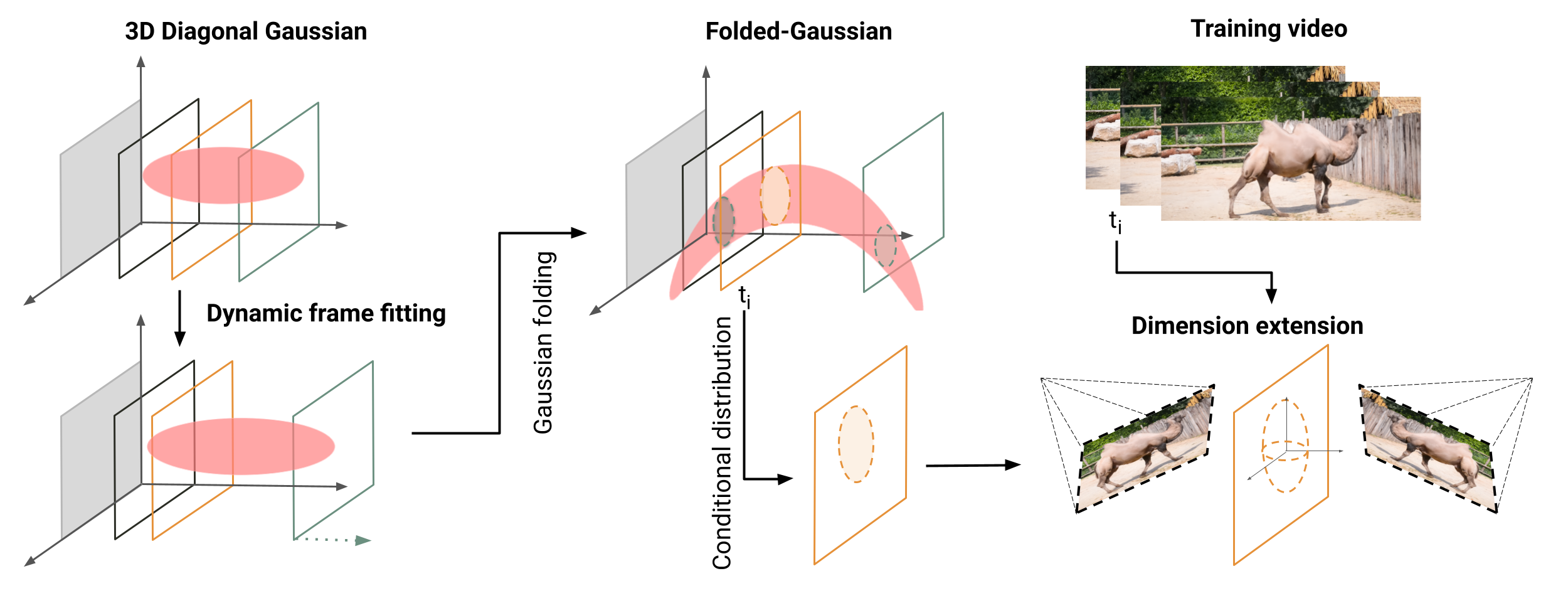}
\captionof{figure}{Graphical summary of our Video Gaussian Splatting (\our{}) model. The initial step involves the use of diagonal 3D Gaussians and frames with equal distances. Then, dynamic frame fitting and Gaussian folding are employed to approximate nonlinear structures within a video stream. Each frame is modeled by 2D Gaussians obtained by conditioning of 3D Folded-Gaussians at frame occurrence time $t_i$. This representation allows for the creation of high-quality renderings of video data and facilitates a wide range of modifications.}

\vspace{1em}
\label{fig:teaser}
}]

\def\thefootnote{*}\footnotetext{These authors contributed equally to this work}\def\thefootnote{\arabic{footnote}}

\begin{abstract}
Implicit Neural Representations (INRs) employ neural networks to approximate discrete data as continuous functions. In the context of video data, such models can be utilized to transform the coordinates of pixel locations along with frame occurrence times (or indices) into RGB color values. Although INRs facilitate effective compression, they are unsuitable for editing purposes. One potential solution is to use a 3D Gaussian Splatting (3DGS) based model, such as the Video Gaussian Representation (VGR), which is capable of encoding video as a multitude of 3D Gaussians and is applicable for numerous video processing operations, including editing. Nevertheless, in this case, the capacity for modification is constrained to a limited set of basic transformations. To address this issue, we introduce the Video Gaussian Splatting (\our{}) model, which enables realistic modifications of video data. To construct \our{}, we propose a novel family of Folded-Gaussian distributions designed to capture nonlinear dynamics in a video stream and model consecutive frames by 2D Gaussians obtained as respective conditional distributions. Our experiments demonstrate that \our{} outperforms state-of-the-art solutions in frame reconstruction tasks and allows realistic modifications of video data. The code is available at: \url{https://github.com/gmum/VeGaS}.
\end{abstract}  

\section{Introduction}
\label{sec:intro}

Implicit Neural Representations (INRs)~\cite{SIREN} employs neural networks to describe discrete data as a smooth, continuous function. They have emerged as a promising method for continuously encoding a variety of signals, including images \cite{klocek2019hypernetwork}, videos \cite{chen2022videoinr}, audio \cite{szatkowski2023hypernetworks}, and 3D shapes \cite{park2019deepsdf}. INRs are frequently utilized in the context of 2D imagery by training networks that map pixel coordinates to RGB color values, thus encoding a structure of images in neural network weights. This approach offers several benefits, including applications in compression \cite{chen2022videoinr}, hyper-resolution \cite{klocek2019hypernetwork}, or as an integral part of generative models \cite{skorokhodov2021adversarial,kania2023hypernerfgan}.

On the other hand, the 3D Gaussian Splatting (3DGS) framework \cite{kerbl20233d}, initially proposed for the modeling of 3D scenes, has recently been adapted with 2D images. In particular, the GaussianImage method \cite{zhang2024gaussianimage} has demonstrated promising results in image reconstruction by efficiently encoding images in the 2D space, with a strong focus on model efficiency and reduced training time. Furthermore, the MiraGe representation \cite{waczynska2024mirage} has demonstrated the feasibility of generating realistic modifications of 2D images.

Similarly to 2D images, INR produces a continuous representation of videos~\cite{chen2022videoinr}. In such a case, neural networks transform the pixel coordinates and time frames into RGB color. Such models provide good reconstruction quality and compression ratios. Unfortunately, INR ultimately failed with the editing of videos.
To solve such a problem, we can use the Gaussian Splatting solution. Video Gaussian Representation (VGR) \cite{sun2024splatter} uses Gaussians in the canonical position and deformation function, which transfers such a Gaussian to each time frame. This model is capable of handling a variety of video processing tasks, such as video editing. Nevertheless, these changes are restricted to linear transformations and translations.

\begin{figure}[htbp]
    \centering
\begin{tabular}{@{}c@{\;}c@{}}
	\rotatebox{90}{\footnotesize Vanishing \quad Original \quad Multiplication \quad Original \quad Scale \quad Original} &
    
    \includegraphics[trim=0 0 0 0,clip,width=0.9\columnwidth]{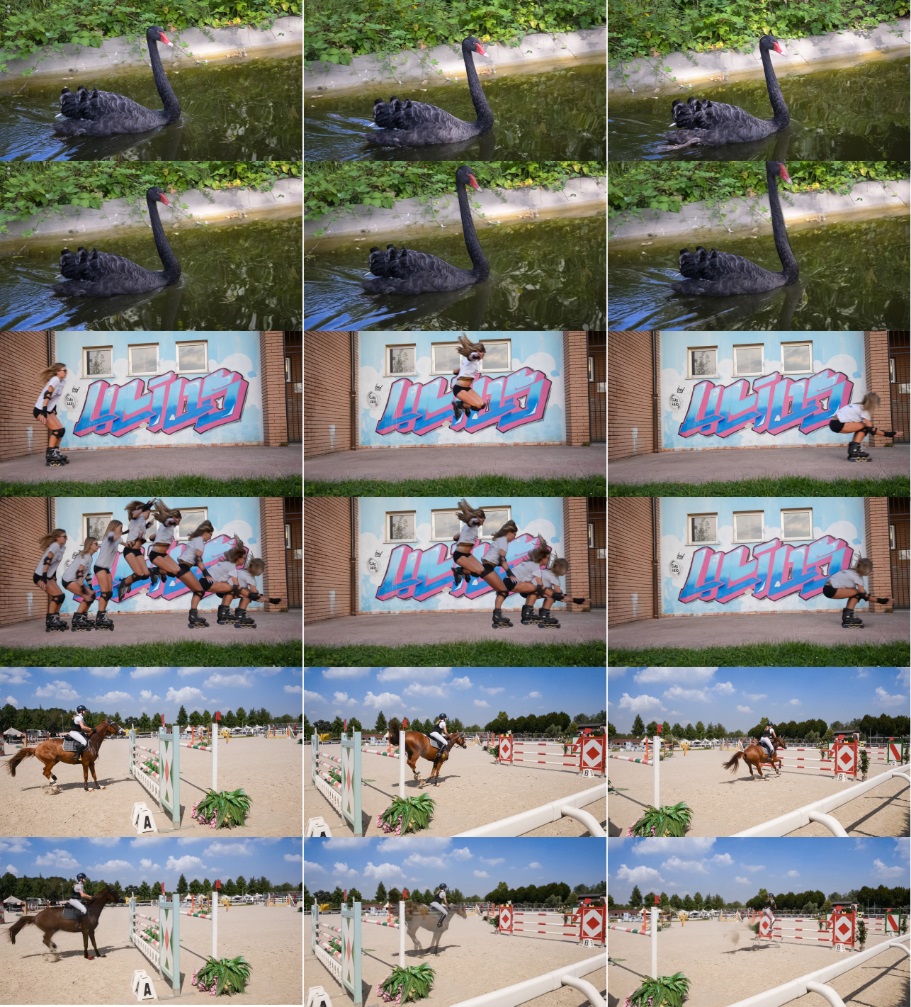}
\end{tabular}
    
    \caption{{\em Video edition}. Note that \our{} enables modification of selected objects on a global scale, including operations such as multiplication and scaling. The model was trained on the DAVIS dataset~\cite{davis}.}
\vspace{-0.5cm}
    \label{fig:sim_1}
\end{figure}

This paper introduces the Video Gaussian Splatting \mbox{(\our{})} model, which presents evidence that the 3DGS approach can be adapted for use with 2D video data. In particular, video frames are treated as parallel planes within a 3D space, and a 3D Gaussian Splatting is employed to model the transitions observed between the content of subsequent frames. By conditioning 3D Gaussian components at specified time points, 2D Gaussians are tailored to the selected frames. It is crucial to emphasize that our solution surpasses the classical Gaussian Splatting model by enhancing its capacity to integrate intricate distributions, thus facilitating more exact modeling of swift alterations in video sequences. In particular, we introduce \ourdens{}, a family of functions that model nonlinear structures and produce classical 2D Gaussian distributions after conditioning. It should be noted that the employment of the 3DGS framework for video modeling allows for the utilization of both extensive Gaussians to represent the background, which remains largely static over time, and brief Gaussians to represent elements present in only a few frames. Furthermore, \our{} employs a MiraGe-based representation to model individual frames, which allows one to modify both the entire video and selected frames, resulting in high-quality renderings, as shown in Figure~\ref{fig:sim_1}.

The following represents a comprehensive account of our significant contributions:
\begin{itemize}
  \item we introduce \ourdens{}, a novel family of functions that model nonlinear structures and can be readily incorporated into the 3D Gaussian Splatting framework,
    \item we propose the \our{} model, which allows for the processing of 2D video data using the \ourdens{},
    \item we conduct experiments that demonstrate the superiority of \our{} for reconstruction tasks and show its efficiency in producing realistic modifications of video data.
\end{itemize}

\section{Related Works}

The decomposition of videos into layered representations enables the utilization of sophisticated video editing techniques. In \cite{kasten2021layered}, the authors decompose an image into textured layers and learn a corresponding deformation field, which allows for efficient video editing. The method outlined in \cite{ye2022deformable} involves the partitioning of videos into discrete motion groups, with each group being driven by an MLP-based representation. INVE~\cite{huang2023inve} employs a bidirectional warping field to facilitate extensive video tracking and editing over extended periods of time. In \cite{chan2023hashing}, the authors propose an improvement to the rendering of lighting and color details. This is achieved by incorporating additional layers and residual color maps, which serve to enhance the representation of illumination effects in the video. CoDeF~\cite{ouyang2024codef} employs a multi-resolution hash grid and a shallow MLP to model frame-by-frame deformations relative to a canonical image. This approach enables editing in the canonical space, with changes effectively propagated across the entire video. A comparable representation is utilized in GenDeF~\cite{wang2023gendef} for the generation of controllable videos. 

The generative potential of latent diffusion models has been harnessed in various research endeavors within the context of video editing \cite{rombach2022high}. In \cite{zhang2023controlvideo}, the authors integrate control signals into the network during video reconstruction, thereby guiding the editing process. A related technique involves frame interpolation to generate edited videos from specifically edited keyframes \cite{ma2024maskint}, while another method employs a token merging approach to incorporate control signals during \cite{li2024vidtome}. Moreover, some works investigate inversion techniques for video editing \cite{li2024video,zhang2024fastvideoedit}.

3D Gaussian Splatting (3DGS)~\cite{kerbl20233d} models 3D static scenes using a set of Gaussian components. Recently, numerous generalizations have been proposed for the representation of dynamic scenes. In \cite{luiten2023dynamictracking}, the authors employ a multiview dynamic dataset coupled with an incremental, frame-based strategy. Nevertheless, this method does not account for inter-frame correlation and requires a considerable amount of storage space for extended sequences. The approach presented in \cite{yang2023deformable3dgs,kratimenos2024dynmf} employs an MLP to represent temporal alterations in Gaussians. In contrast, in \cite{wu20234dgaussians} the authors utilize an MLP in conjunction with a decomposed neural voxel encoding technique to enhance training and storage effectiveness. In \cite{liang2023gaufre}, dynamic scenes are divided into dynamic and static segments, which are optimized independently and then combined to facilitate decoupling. Other research has sought to enhance the reconstruction of dynamic scenes by incorporating external priors. For instance, diffusion priors have been shown to serve as effective regularization terms in the optimization process \cite{zhang2024bags}. In \cite{duan20244d}, the authors propose 4DRotorGS which employs a four-dimensional Gaussian, with the fourth dimension dedicated to time.

\begin{figure*}[!h]
    \centering
    \includegraphics[trim=0 0 0 0,clip,width=0.22\textwidth]{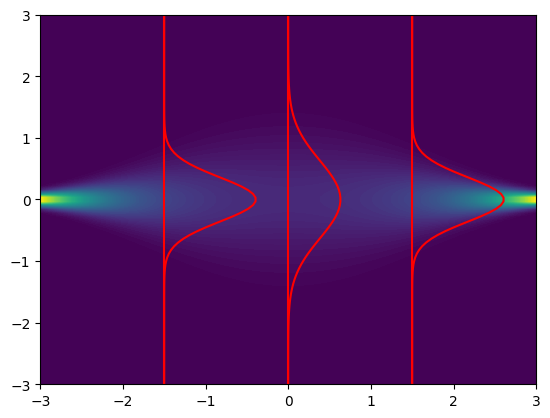}
    \includegraphics[trim=0 0 0 0,clip,width=0.22\textwidth]{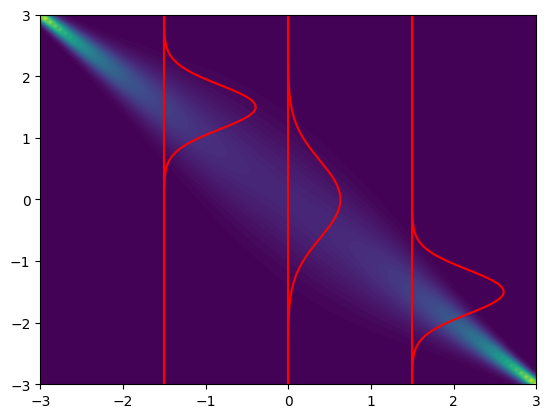}
    \includegraphics[trim=0 0 0 0,clip,width=0.22\textwidth]{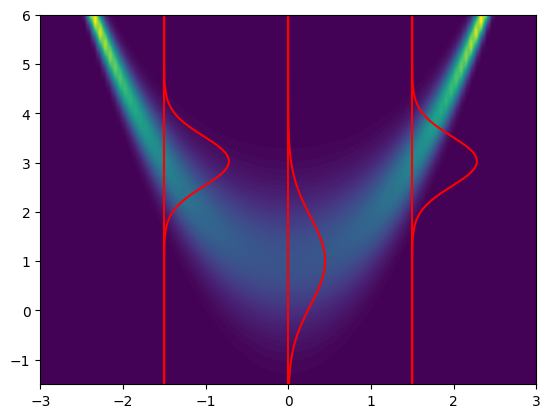}
    \includegraphics[trim=0 0 0 0,clip,width=0.22\textwidth]{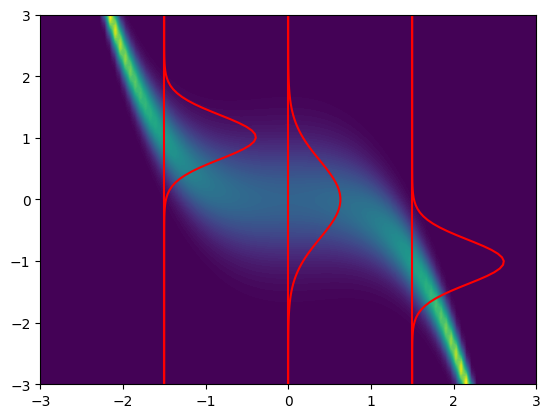}\\
    $f(x) = 0$ \quad\quad\qquad\qquad $f(x) = -x$ \quad\quad\qquad\qquad $f(x) = x^2$ \quad\quad\qquad\qquad $f(x) = x^3$\\ 
    \caption{Folded-Gaussian distribution is capable of capturing both linear and nonlinear patterns. It is crucial to highlight that the conditional distributions (marked in red) are classical Gaussians.}

    \label{fig:2D_dist}
\end{figure*}

Furthermore, 3DGS was utilized to modify scene geometry based on the underlying meshes. In \cite{gao2024meshbased}, the positioning of 3D Gaussians on an explicit mesh enables the utilization of mesh rendering to facilitate adaptive refinement. This method is contingent upon the use of an extracted mesh as a proxy; therefore, it is inoperable in the event of a failure in the mesh extraction process. On the other hand, in \cite{guedon2023sugar} explicit meshes are derived from 3DGS representations through the regularization of Gaussians over surfaces. This process, which involves a significant optimization and refinement phase, is particularly resource-intensive. Another example is given in \cite{huang2023sc-gs}, where sparse control points are used for 3D scene dynamics. However, this approach encounters difficulties with extensive edit movements and requires precise static node selection. In turn, GaMeS~\cite{waczynska2024games} integrates 3DGS with mesh extraction, though this approach is only effective for static scenes. In contrast, D-MiSo~\cite{waczynska2024d} is a mesh-based approach, specifically designed for dynamic scenes, which employs a simple 3DGS technique pipeline to allow real-time editing of dynamic scenes.

In \cite{sun2024splatter}, the authors introduce the Video Gaussian Representation (VGR), which employs 3D Gaussian Splatting to model video data. This approach is closely related to ours and therefore represents the most reasonable baseline for the \our{} model. VGR uses Gaussians in the canonical position, transferring them further to each frame occurrence time, and a deformation function. This model is capable of handling a variety of video processing tasks, such as video editing. Nevertheless, the possible changes are restricted to linear transformations and translations, which represents a limitation relative to our proposed solution.

\section{\ourdens{}}

In this section, we introduce the concept of a Folded-Gaussian distribution, which can be seen as a generalization of a classical Gaussian distribution in order to capture nonlinear structures. It should be noted that \ourdens{} constitute a novel family of distributions that we further employ to represent video data (see the next section). Accordingly, in order to emphasize this relationship, a terminology based on the concept of a space-time variable is employed, given that each video can be regarded as a sequence of successive frames occurring at discrete time points. For the reader's convenience, we begin with a simple two-dimensional toy example, before extending our presentation to the multidimensional case.

\paragraph{Toy Example in $\R^2$}

Our toy example starts with a two-dimensional Gaussian distribution $\N(\m,\Sigma)$ of a space-time random variable $\x=(s,t)\in \R\times \R$, which is given by a mean vector $\m=(m_s, m_t)$ and a covariance matrix 
\begin{equation}\label{eq:sigma_2D}
\Sigma=
\begin{bmatrix}
\sigma_s^2 & 0 \\
0 & \sigma_{t}^2
\end{bmatrix}.
\end{equation}
In this case, the density function is defined by the following formula:
\begin{equation}\label{eq:pdf_2d}
N(\m,\Sigma)( \x ) = N(m_s,\sigma_s^2)(s) \cdot N(m_t,\sigma_t^2)(t),
\end{equation}
where
\begin{equation}\label{eq:normal_pdf}
N(m,\sigma^2)(x)=
\frac{1}{\sqrt{2 \pi} \sigma}
\exp\left(-\frac{|x-m |^2}{2 \sigma^2} \right).
\end{equation}
Using such a distribution, we can model ellipses, which can be considered as a simple linear structure spanned along the coordinate axes. Therefore, we propose a generalization of a classical 2D Gaussian that allows us to deal with nonlinear patterns. Specifically, we are looking for a two-dimensional distribution for which conditioning on the time variable would produce one-dimensional Gaussians aligned along an arbitrary curve (not necessarily linear), as shown in Figure~\ref{fig:2D_dist}. A possible solution is to ensure that a conditional distribution of $s|t$ is a Gaussian distribution
\begin{equation}\label{eq:cond_pdf_2d}
\N(m_s+f(m_t-t), a(t) \sigma_s^2), 
\end{equation}
where $f\colon \R\to \R$ and $a\colon \R\to [0,1]$ are functions designed to capture a desired time-dependent shift and rescaling of the space variable, respectively. In practice, we use a polynomial of a given order as $f$ and the likelihood of the time variable (scaled to the unit interval) as $a$, i.e.,
\begin{equation}\label{eq:a_func_2d}
a(t)=\frac{\N(m_t,\sigma^2_t)(t)}{\N(m_t,\sigma^2_t)(m_t)}.
\end{equation}
Finally, to recover a joint distribution of a space-time variable, we can apply the standard chain rule for random variables, which leads to a density function given by the following formula: \begin{equation}\label{eqpdf_gauss_2d}
\N(m_s+f(m_t-t), a(t) \sigma_s^2)(s|t)\cdot \mathcal{N}(m_t,\sigma_t^2)(t).
\end{equation}
It is important to note that while marginal distributions of $t$ and conditional distributions of $s|t$ are both Gaussian, the resulting joint distribution is not a Gaussian anymore. This makes it an interesting object not only for applications, but also for further theoretical studies. To ensure the applicability of our contribution to a wider range of contexts, we extend the discussion to an arbitrary dimension in the following paragraph.

\paragraph{\ourdens{} in $\R^d$}
We begin with a multivariate Gaussian distribution $\mathcal{N}(\m,\Sigma)$, which is defined for a space-time random variable $\x=(\s,t)\in \R^{d-1}\times \R$ by a mean vector $\m=(\m_\s,m_t)$ and a covariance matrix
\begin{equation}\label{eq:cov}
\Sigma=
\begin{bmatrix}
\Sigma_{\s} & 0 \\
0 & \sigma_{t}^2
\end{bmatrix},
\end{equation}
where $\Sigma_\s$ denotes the diagonal covariance matrix of the space component of $\x$. The probability density function (PDF) is then factorized into two independent normal densities, i.e., 
\begin{equation}\label{eq:gauss_2d}
\mathcal{N}(\m,\Sigma)( \x ) = \mathcal{N}({\m}_{\s},\Sigma_\s)(\s) \cdot \mathcal{N}(m_t,\sigma_t^2)(t).
\end{equation}
Note that such a distribution permits the modeling of only simple linear structures that are spanned along coordinate axes (see the leftmost picture in Figure~\ref{fig:2D_dist} for an appropriate example in the two-dimensional case).

To address the aforementioned limitation, we propose incorporating non-trivial conditioning, which allows for a more flexible representation. In particular, we apply the following time-dependent transformation on the space variable:
\begin{equation}\label{eq:transform_s}
\s \to \sqrt{a(t)}(\s-\m_\s)+\m_\s+f(m_t-t),
\end{equation}
where $a\colon \R\to (0,1]$ and $f\colon \R\to \R^{d-1}$ are suitably chosen functions. This yields the new space variable with the conditional normal density
\begin{equation}\label{eq:cond_dens}
\begin{split}
\N(\m_{s|t},\Sigma_{s|t},& a, f)(\s|t)= \\
&\mathcal{N}(\m_\s+f(m_t-t),a(t)\Sigma_\s) (\s|t),
\end{split}
\end{equation}
which in turn gives rise to a novel Folded-Gaussian distribution with the PDF defined as\footnote{Note that in this case we are using the standard chain rule for random variables.}:
\begin{equation}\label{eq:foldgauss}
\begin{split}
\mathcal{FN}(\m,&\Sigma, a,f)(\x) = \\
& \N(\m_{s|t},\Sigma_{s|t},a,f)(s|t) \cdot \mathcal{N}(m_t,\sigma_t^2)(t),
\end{split}
\end{equation}
where $\s|t$ represents the new time-conditioned space random variable and $\x=(\s|t,t)$ (we do not change the notation as we do not believe it would cause confusion). A notable benefit of \ourdens{} is their ability to effectively capture a range of relationships present in the data. This is due to the inherent flexibility in selecting the functions $f$ and $a$. In the context of the \our{} model, the polynomial function $f$ (with trained coefficients) is employed in conjunction with the likelihood-based function $a$ (as in Equation~\ref{eq:a_func_2d}). Consequently, our approach allows us to encompass both linear and non-linear patterns, as illustrated in Figure~\ref{fig:2D_dist}, which refers to a simplified case of two-dimensional distributions. Furthermore, the incorporation of likelihood-based time-dependent rescaling leads to the disappearance of the tails of \ourdens{}, thus facilitating the capture of elements present in only a portion of a video stream (an optimal scenario would entail these elements initially approaching the camera and subsequently receding from view).

The following paragraph offers further theoretical insight into the Folded-Gaussian distribution, including the formal arguments that underpin Equations \eqref{eq:cond_dens} and \eqref{eq:foldgauss}.

\paragraph{Theoretical Study}
It should first be noted that the transformation given in Equation \eqref{eq:transform_s} is of an affine form $A\s + b$. Consequently, the distribution of the random variable $s|t$ is also Gaussian with parameters $A\m_s + b$ and $A\Sigma_sA^T$. Given that in our case we have 
\begin{equation}\label{eq:Ab_formulas}
A=\sqrt{a(t)}I_d, \;\mathrm{b}=\left(1-\sqrt{a(t)}\right)\m_\s + f(m_t-t),
\end{equation}
we can conclude that
\begin{equation}\label{eq:cond_dens_s|t}
\s|t\sim \mathcal{N}(\m_\s+f(m_t-t),a(t)\Sigma_\s),
\end{equation}
which justifies the assertion made in Equation~\eqref{eq:cond_dens}.
Moreover, a straightforward assessment of the correctness of the definition of the Folded-Gaussian distribution, as given by Equation~\eqref{eq:foldgauss}, can be conducted through the following calculation:
\begin{equation}\label{eq:integral_proof}
\begin{split}
& \!\!\!\!\!\int\mathcal{FN}(\m,\Sigma,a,f)(\x) d  x  = \\ & \!\!\!\!\!\int\left(\int\N(\m_{s|t},\Sigma_{s|t},a,f)(\s|t) d  s|t\right)  \mathcal{N}(m_t,\sigma_t^2)(t) d  t\\
& \!\!\!\!\!= \int \mathcal{N}(m_t,\sigma_t^2)(t) d  t = 1.
\end{split}
\end{equation}
Similarly, based on Equation~\eqref{eq:cond_dens}, the PDF of the conditional distribution of the random variable $\s|t$ can be computed as follows:
\begin{equation}\label{eq:cond_proof}
\begin{split}& \frac{\mathcal{FN}(\m,\Sigma,a,f)(\s|t,t)}{\int\mathcal{FN}(\m,\Sigma,a,f)(\s|t,t) d  s}  = \\ & \frac{ \N(m_t,\sigma^2_t)(t)\cdot\N(\m_{s|t},\Sigma_{s|t},a,f)(\s|t)}{\mathcal{N}(m_t,\sigma_t^2)(t)\cdot \int\N(\m_{s|t},\Sigma_{s|t},a,f)(\s|t) d  s|t} \\
& = \N(\m_{s|t},\Sigma_{s|t},a,f)(s|t),
\end{split}
\end{equation}
which corroborates our previous assertion.

\section{Video Gaussian Splatting}

This section introduces our Video Gaussian Splatting \mbox{(\our{})} model. We start with a brief overview of the 3D Gaussian Splatting (3DGS) \cite{kerbl20233d} method and then proceed to the MiraGe \cite{waczynska2024mirage} approach for 2D images, while at the same time tailoring the presentation for straightforward integration into our proposed solution. The section concludes with a detailed description of the \our{} model.

\paragraph{3D Gaussian Splatting}

The 3D Gaussian Splatting (3DGS)~\cite{kerbl20233d} method employs a family of three-dimensional Gaussian distributions 
\begin{equation}\label{eq:3dgs}
\G_{\text{3DGS}} = \{ (\N(\m,\Sigma), \rho, c) \},
\end{equation}
characterized by a set of attributes, including location (mean) $\m$, covariance matrix $\Sigma$, opacity $\rho$, and color $c$. In practice, the covariance matrix $\Sigma$ is factorized as
$
\Sigma=RSSR^T,
$
where $R$ is the rotation matrix and $S$ is a diagonal matrix containing the scaling parameters. Therefore, it is also possible to use the notation $N(\m,R,S)$ instead of $N(\m, \Sigma)$.

The efficiency of the 3DGS technique is primarily attributable to its rendering process, which involves the projection of 3D Gaussians onto a two-dimensional space.
Throughout the training process, all parameters are optimized according to the mean square error (MSE) cost function. Since such a procedure often results in local minima, 3DGS can employ supplementary training methods that include component creation, removal, and repositioning based on the proposed heuristic, which is both a fast and effective strategy. In addition, the GS training process is executed within the CUDA kernel, allowing for rapid training and real-time rendering.

\paragraph{Gaussian Spatting for 2D images}

The MiraGe \cite{waczynska2024mirage} approach employs the 3DGS technique to accommodate 2D images. This is accomplished by employing flat Gaussians positioned on the plane spanned by the canonical vectors $\e_1=(1,0,0)$ and $\e_2=(0,1,0)$, which gives rise to a specific type of parametrization. In essence, this method deals with a family of 3D Gaussian components of the form
\begin{equation}\label{eq:mirage}
\G_{\text{MiraGe}} = \{(\N(\m, R, S), \rho, c )\}, 
\end{equation}
where $\m=(m_1,m_2,0)$, $S=\mathrm{diag}(s_1,s_2,\varepsilon)$, and
\begin{equation}
R =[\r_1,\r_2,\e_3]= \begin{bmatrix}
\cos{\theta} & -\sin{\theta} & 0 \\
\sin{\theta} & \cos{\theta} & 0\\
0 & 0 & 1 
\end{bmatrix}.
\end{equation}
($\varepsilon$ is a small positive constant used to ensure compatibility with the three-dimensional framework.) Subsequently, utilizing the parametrization proposed by the GaMeS~\cite{waczynska2024games} model, such flat Gaussians can be represented by three points (triangle face)
\begin{equation}\label{eq:traingle_face}
    V=[\m,\v_1,\v_2]=\ga(\m,R,S),
\end{equation}
with the vertices defined as
$\v_1 = \m + s_1 \r_1$, and $\v_2 = \m + s_2 \r_2$. On the other hand, given a face representation $V=[\m,\v_1,\v_2]$, the Gaussian component
\begin{equation}\label{eq:gausscomp}
    \N(\m, R, S)=\N(\ga^{-1}(V))
\end{equation}
can be reconstructed through the mean $\m$, the rotation matrix $R=[\r_1,\r_2,\e_3]$, and the scaling matrix $S = \mathrm{diag}(s_1,s_2,\varepsilon)$, where the parameters are defined by the following formulas: 
\begin{equation}\label{eq:gaussrecons1}
    \r_1  =  \frac{\v_1-\m}{\| \v_1-\m \|},\;\; \r_2 = \mathrm{orth}(\v_2-\m;\r_1,\e_2),
\end{equation}
\begin{equation}\label{eq:gaussrecons2}
    \s_1 = \|\v_1-\m\|, \;\; s_2 = \langle \v_2-\m, \r_2 \rangle.
\end{equation}
In this context, $\mathrm{orth}(\cdot)$ represents one iteration of the Gram-Schmidt process~\citep{bjorck1994numerics}. We would like to highlight that the formulas presented above have been adjusted to align with our framework and may therefore differ slightly from those provided in \cite{waczynska2024games,waczynska2024mirage}.

The use of the GaMeS parameterization allows for the modification of the position, scale, and rotation of Gaussians by altering the underlying triangle face. Furthermore, the MiraGe extension facilitates the manipulation of 2D images within 3D space, thereby creating the illusion of three-dimensional effects. 

\begin{figure}[htbp]
    \centering
    \includegraphics[trim=0 0 0 0,clip,width=\columnwidth]{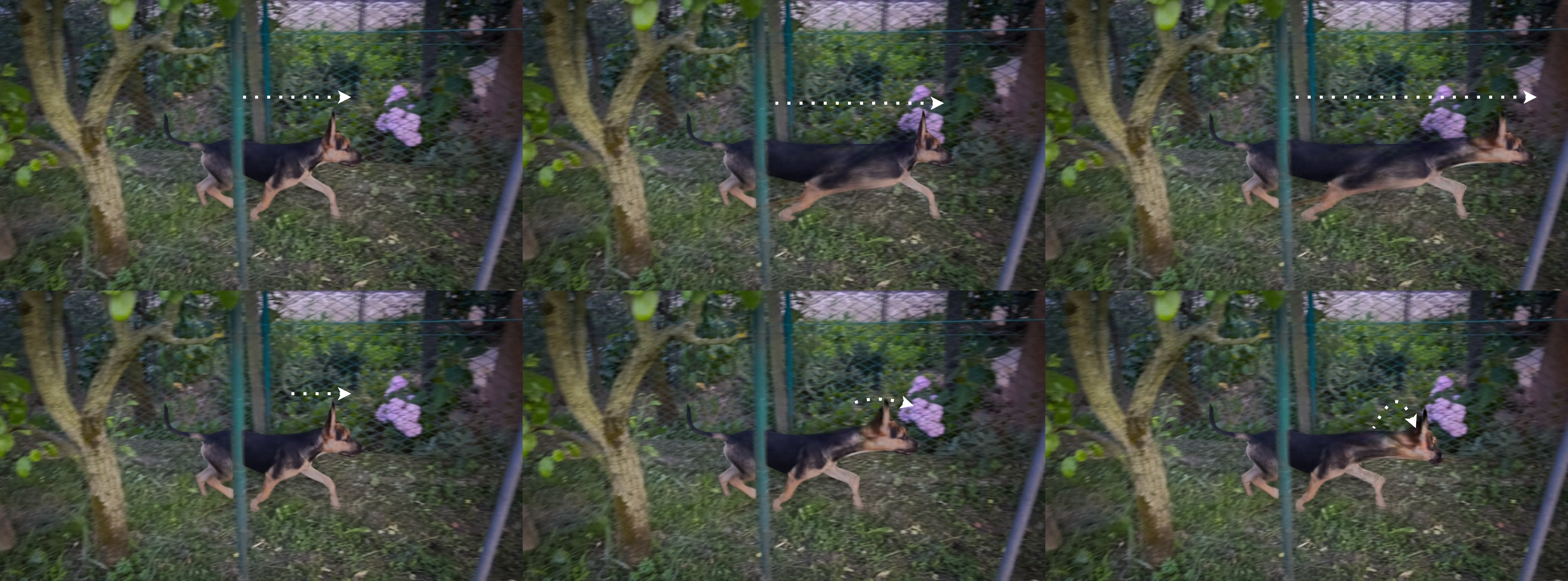}
    \caption{{\em Video edition}. Note that \our{} permits selection of a single frame and modification of some of its elements. The model was trained on the DAVIS dataset~\cite{davis}.}
\vspace{-0.5cm}
    \label{fig:sim_2}
\end{figure}

\paragraph{Video Gaussian Splatting}

Consider a video comprising a sequence of frames 
$[I_{t_1}, \ldots, I_{t_n}]$,
indexed by their occurrence times scaled to the unit interval $[0,1]$. In this context, the MiraGe model may be employed for each consecutive frame, as it can be treated as a separate 2D image. Consequently, this would result in a joined family of 3D Gaussian distributions \begin{equation}\label{eq:miragevideo}
\G_{\text{MiraGe}}^{t_1}\cup \ldots \cup \G_{\text{MiraGe}}^{t_n},
\end{equation}
where each $\G_{\text{MiraGe}}^{t_i}$ is given by Equation~\eqref{eq:mirage}, which could be considered an adequate representation of the entire data. However, such an approach completely ignores the relationships that naturally exist within a video stream.

To overcome the mentioned limitation, we propose to construct Gaussians related to successive frames by conditioning of corresponding three-dimensional Folded-Gaussian distributions at frames occurrence times. The resulting Video Gaussian Splatting (\our{}) model is thus formally defined as a collection on 3D \ourdens{}
\begin{equation}\label{eq:vegas_equiv}
\G_{\text{\our{}}} = \{ (\mathcal{FN}(\m, \Sigma, a,f), \rho, c) \},
\end{equation}
where for each component, three-dimensional extensions (see the preceding paragraph) around two-dimensional conditional distributions \begin{equation}\label{eq:vegas_equiv_cond}
\N(\m_{s|t_1},\Sigma_{s|t_1}, a, f), \ldots , \N(\m_{s|t_n},\Sigma_{s|t_n}, a, f)
\end{equation} 
are built using Equation~\eqref{eq:cond_dens}, with respect to the common opacity $\rho$ and color $c$.

It is important to highlight our method does not use the fixed frames occurrence times $t_1, t_2,\ldots, t_n$, but learns them through an optimization procedure that guarantees superior reconstruction quality. Specifically, we use the dynamic frame fitting function $f_t\colon \mathbb{Z}_{+} \to [0,1]$, which maps the frame number $k$ to its scaled occurrence time $t_k$ as follows
\begin{equation}\label{eq:frame_fitting}
t_k = f_t(k) = \sum_{i=1}^{k} \sigma{(w)}_i = \sum_{i=1}^{k}\frac{e^{w_i}}{\sum_{j=1}^{n-1}e^{w_j}},
\end{equation}
where $w_1, w_2,..., w_{n-1}$ are trainable parameters. (For interpolated frames between $I_k$ and $I_{k+1}$, we use evenly spaced times between $t_k$ and $t_{k+1}$).
Additionally, it should be noted that in the VeGaS model we are dealing with two types of spatial distributions. First, we have \ourdens{}, which represent the dynamics in the video stream. Second, we generate flat conditional distributions, which are then extended to 3D Gaussians by adding a small orthogonal component (controlled by the value of $\varepsilon$). In this case, two opposing cameras reconstruct a single 2D image -- one produces the original image, while the other produces its mirrored version.

\begin{table*}
\footnotesize
    \centering
    \caption{{\em Frame reconstruction.} Performance of the \our{} model on the evaluation setting proposed in \cite{sun2024splatter}, using various videos from the DAVIS dataset~\cite{davis}, in terms of the PSNR metric. Note that, in each situation, \our{} obtains the best metric scores.}
    \label{table:sota-comparison}
    \begin{tabular}{lcccccccc}
        \toprule
        \textbf{Model} & Bear & Cows & Elephant & Breakdance-Flare & Train & Camel & Kite-surf & Average
 \\
 \midrule
 Omnimotion \cite{wang2023tracking} & 22.96 & 23.93 & 26.59 & 24.45 & 22.85 & 23.98 & 23.72 & 24.07\\
CoDeF~\cite{ouyang2024codef} & 29.17 & 28.82 & 30.50 & 25.99 & 26.53 & 26.10 & 27.17 & 27.75 \\
VGR \cite{sun2024splatter} & 30.17 & 28.24 & 29.82 & 27.18 & 28.09 & 27.74 & 27.82 & 28.44 \\
\our{}-Full (our)  &  \textbf{31.79} &  27.64 & \textbf{30.93} & \textbf{29.37} & \textbf{31.20} &  \textbf{30.76} & \textbf{35.84} & \textbf{31.08}\\
\our{}-480p (our)   &  \textbf{33.23} & \textbf{30.27} & \textbf{33.28} & \textbf{33.19} & \textbf{32.86} & \textbf{32.23}  &  \textbf{38.23} & \textbf{33.31}\\

        \bottomrule
    \end{tabular}
\end{table*}

\section{Experiments}

This section presents an extensive experimental study of the \our{} model in a variety of settings, which compares its efficiency with respect to a diverse range of state-of-the-art solutions.

\begin{figure*}[htbp]
    \centering
    \includegraphics[trim=0 0 0 0,clip,width=0.85\textwidth]{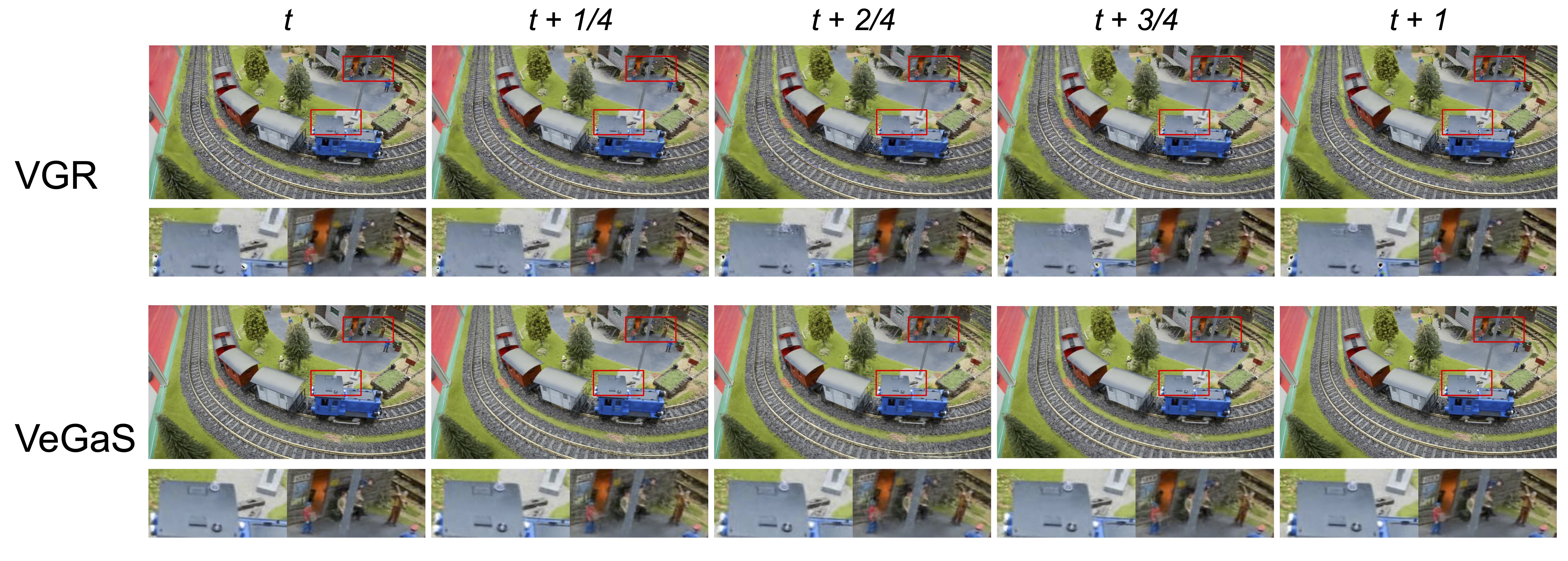}
    \caption{{\em Frame interpolation.} Qualitative results obtained by \our{} and VGR~\cite{sun2024splatter} on a selected video object from the DAVIS dataset~\cite{davis}. Frames at times $t$ and $t+1$ are reconstructions of two consecutive original frames, while frames at times $t+1/4$, $t+2/4$, and $t+3/4$ are interpolated. Note that \our{} produces outcomes that are slightly more favorable.}\vskip-0.3cm
    \label{fig:interp}
\end{figure*}

\begin{table*}
\footnotesize
  \centering
  \caption{{\em Frame reconstruction.} Performance of the \our{} model on setting proposed in \cite{zhao2023dnerv}, using various videos from the DAVIS dataset~\cite{davis}, in terms of the PSNR and SSIM metrics. Note that \our{} outperforms all baseline models. }
  \begin{tabular}{lcccccccccc}
 \toprule
       & \multicolumn{2}{c}{NeRV \cite{chen2021nerv}} & \multicolumn{2}{c}{E-NeRV \cite{li2022nerv} } & \multicolumn{2}{c}{HNeRV \cite{chen2023hnerv} } & \multicolumn{2}{c}{DNeRV \cite{zhao2023dnerv}} & \multicolumn{2}{c}{\our{} (our)} \\
       & PSNR$\uparrow$ & SSIM$\uparrow$ & PSNR$\uparrow$ & SSIM$\uparrow$ & PSNR$\uparrow$ & SSIM$\uparrow$ & PSNR$\uparrow$ & SSIM$\uparrow$ & PSNR$\uparrow$ & SSIM$\uparrow$ \\
    \midrule
    Blackswan &28.48 & 0.812 & 29.38 & 0.867  &30.35 & 0.891 & 30.92 & 0.913 &\textbf{34.92} & \textbf{0.932} \\
   Bmx-bumps  & 29.42 & 0.864 & 28.90 & 0.851  & 29.98 & 0.872 & 30.59 & 0.890 &\textbf{33.01} & \textbf{0.915} \\
   Bmx-trees  & 26.24 & 0.789 & 27.26 & 0.876  & 28.76 & 0.861 & 29.63 & 0.882 & \textbf{31.78} & \textbf{0.896} \\
   Breakdance &26.45 & 0.915 &28.33 & 0.941  &30.45 & 0.961 & 30.88 & \textbf{0.968} &\textbf{32.27} & 0.950  \\
    Camel	  & 24.81 & 0.781 & 25.85 & 0.844  & 26.71 & 0.844 & 27.38 & \textbf{0.887} &\textbf{31.12} & 0.886  \\
   Car-round  & 24.68  & 0.857 & 26.01 & 0.912  & 27.75 & 0.912 & 29.35 & 0.937 &\textbf{32.75} & \textbf{0.941} \\
   Car-shadow &26.41 & 0.871 &30.41 & 0.922  &31.32 & 0.936 & 31.95 & 0.944 &\textbf{36.41} & \textbf{0.956}  \\
   Car-turn   & 27.45 & 0.813 & 29.02 & 0.888  & 29.65 & 0.879 & 30.25 & \textbf{0.892} & \textbf{31.44} & 0.852 \\
   Cows       & 22.55 & 0.702 & 23.74 & 0.819  &24.11 & 0.792 &24.88 & 0.827 &\textbf{27.97} & \textbf{0.834} 	\\
   Dance-twril&25.79 & 0.797 &27.07 & 0.864  &28.19 & 0.845 &29.13 & \textbf{0.870} &\textbf{30.45} & 0.850 	\\
   Dog        &28.17 & 0.795 &30.40 & 0.882  &30.96 & 0.898 & 31.32 & 0.905 & \textbf{34.52} & \textbf{0.914} 		\\
  \midrule
   Average     &26.40 & 0.818 & 27.85 & 0.879 & 28.93 & 0.881 & 29.66 & 0.901 &\textbf{32.42} & \textbf{0.902} \\
   \bottomrule
  \end{tabular}
  \label{vr_davis}
\end{table*}

\begin{table*}
\footnotesize
    \centering
    \caption{{\em Ablation study.} Effect of a batch size and a degree of the polynomial function $f$ on the performance of \our{} in terms of the PSNR metric and the final number of Gaussians (in parentheses). The model was trained on the Bunny dataset \cite{bunny}.}
    \label{table:bs-pol}
    \begin{tabular}{lcccccc}
        \toprule
           \multicolumn{1}{c}{} & \multicolumn{5}{c}{Polynomial degree} & \multicolumn{1}{c}{}  \\
            Batch size  & 1 & 3 & 5 & 7 & 9 & Mean training time \\
            \midrule
            1     & 36.73 (1.26M) & 37.31 (1.77M) & 37.30 (1.73M) & 37.36 (1.72M) & 37.42 (1.83M) & 31m20s\\
            3   & 38.15 (0.57M) & 38.31 (0.58M) & 38.39 (0.59M) & \textbf{38.53 (0.62M)} & 38.24 (0.59M) & 56m30s\\
            5    & 37.84 (0.33M) & 37.94 (0.32M) & 37.95 (0.32M) & 37.92 (0.31M)  & 37.94 (0.31M) & 1h15m29s \\
            \bottomrule
        \end{tabular}
        \vspace{-0.3cm}
\end{table*}

\paragraph{Datasets}
The efficacy of our method was evaluated on two datasets: the Bunny dataset~\cite{bunny} and the DAVIS dataset~\cite{davis}. The Bunny dataset comprises 132 frames with a resolution of 720$\times$1280. In accordance with the specifications outlined in \cite{chen2023hnerv}, the video is cropped to a resolution of 640$\times$1280. The DAVIS dataset is a high-quality and high-resolution collection of videos utilized for the purpose of video object segmentation. It encompasses a multitude of videos, each comprising a total of less than 100 frames. This dataset is accessible in two distinct versions: a full-resolution version and a more compact 480p version.

\paragraph{Implementation Details}
The initialization process entails the uniform sampling of Gaussian means $m_1$ and $m_2$, which are positioned as points within a two-dimensional bounding box. The activation function utilized for $m_t$ is the sigmoid function, which is initialized in a manner that ensures the resulting values from the activation process are uniformly distributed between $0$ and $1$. Similarly, the exponential function is employed as the activation function for $\sigma_t$, with the initial values distributed uniformly between $0.01$ and $1$. The coefficients of the polynomial function $f$ are sampled uniformly between $-1$ and $1$. Furthermore, the rotation matrix is parameterized as a single number (rotation angle) and the activation function employed is sigmoid multiplied by $2\pi$. The angle is initialized to be uniformly distributed between $0$ and $2\pi$.

The model is trained for 30,000 steps at a batch size of 3, utilizing a polynomial function of degree 7 and 500,000 initial Gaussians, unless otherwise specified.
The learning rates, densification, pruning, and opacity reset settings are all consistent with the 3DGS~\cite{kerbl20233d} framework. In accordance with the MiraGe approach, two cameras are utilized: the initial camera generates the original image, while the second camera produces its mirrored version.

\paragraph{Frame Reconstruction}

\begin{figure}[htbp]
    \centering
    \includegraphics[trim=0 0 0 0,clip,width=\columnwidth]{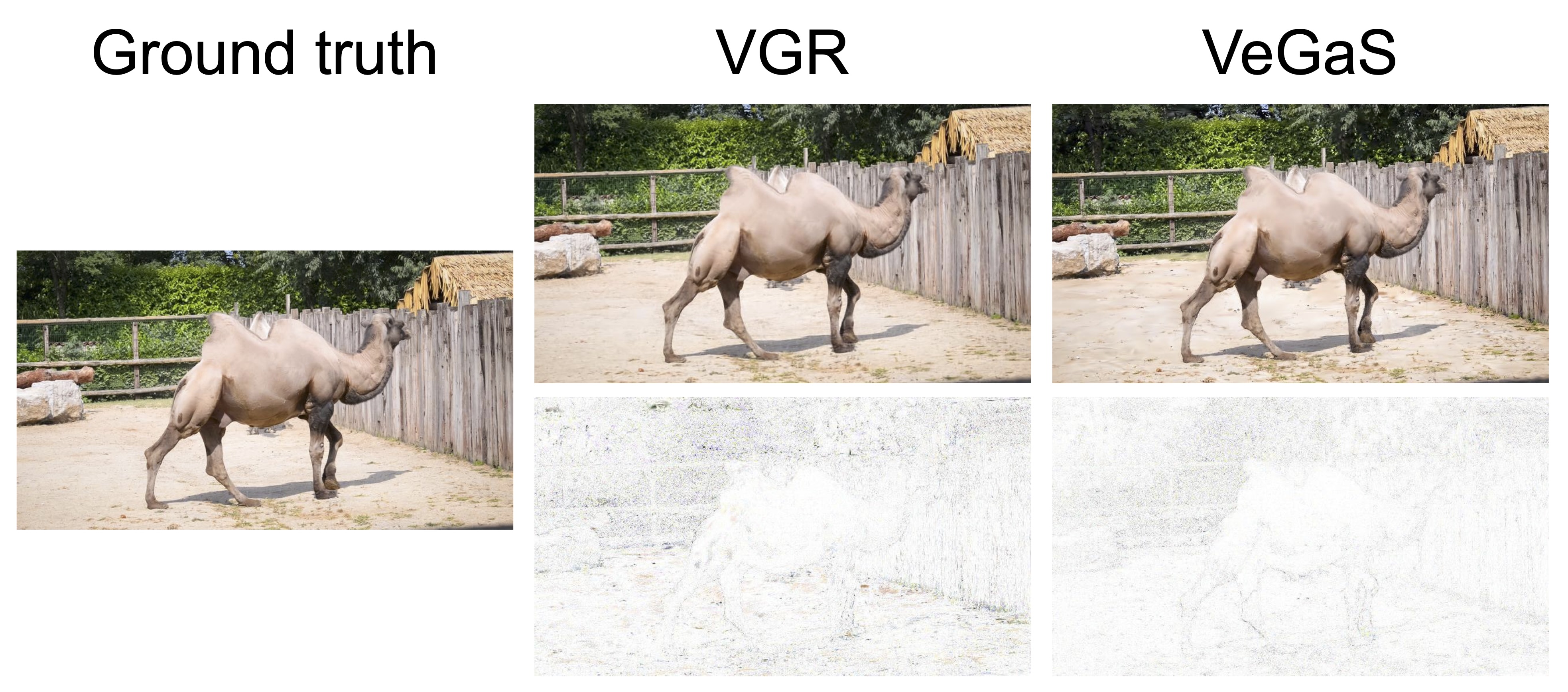}
    \caption{{Frame reconstruction.} Qualitative results obtained by \our{} and VGR~\cite{sun2024splatter} on a selected video object from the DAVIS dataset~\cite{davis}. The bottom row illustrates the discrepancy between the ground truth and the reconstructed image. Note that \our{} is capable of reconstructing videos with high quality and fidelity.}
\vspace{-0,2cm}
    \label{fig:diff}
\end{figure}

We conducted a series of experiments to assess the efficacy of our method in frame reconstruction tasks. The first experimental setup was adapted from that proposed in \cite{sun2024splatter}, where the authors introduce the VGR model and evaluate its performance compared to two state-of-the-art baselines, namely Omnimotion \cite{wang2023tracking} and CoDeF~\cite{ouyang2024codef}. Table \ref{table:sota-comparison} presents the values of rendering quality metrics for various videos from the DAVIS dataset. As there is no information in \cite{sun2024splatter} on the resolution used for the evaluation, we report our results on both cases. It should be noted that, in each situation, \our{} obtains the best metric scores. Furthermore, our model is capable of reconstructing videos with high quality and fidelity, as illustrated in Figure~\ref{fig:diff}, which provides visualizations based on a selected object from the DAVIS dataset~\cite{davis}.

In the other experiments, the scene setting proposed by the authors of \cite{zhao2023dnerv} was used to evaluate the DNeRV model against various NeRF-based baselines, namely \cite{chen2021nerv}, E-NeRV \cite{li2022nerv}, HNeRV \cite{chen2023hnerv}, and DNeRV \cite{zhao2023dnerv}. In accordance with this protocol, the full-resolution version of each scene is center-cropped to a resolution of 960$\times$1920. The results (i.e., the values of the rendering quality metrics) are presented in Table~\ref{vr_davis}, which shows the values obtained for various video objects from the DAVIS dataset. It should be noted that \our{} outperforms all the considered NeRF-based models.

\begin{table}
\footnotesize
    \centering
    \caption{{\em Ablation study.} Effect of the initial number of Gaussians on the performance of \our{} in terms of the PSNR metric, final number of Gaussians, and training time. The model was trained on the Bunny dataset \cite{bunny} with batch size 3 and polynomial degree 7.}
    \label{table:init-gauss}
    \begin{tabular}{lccc}
        \toprule
            Initial Gaussians  & PSNR↑ & Final Gaussians & Training time \\
            \midrule
            0.10M     & 38.53 & 0.62M & 57m00s\\
            0.20M    & 38.85 & 0.62M & 57m29s\\
            0.30M    & 38.99 & 0.62M & 58m03s\\
            0.40M      & 38.96 & 0.64M & 59m36s  \\
            0.50M      & \textbf{39.02} & 0.65M & 58m58s  \\
            0.60M      & 38.86 & 0.66M & 1h00m53s  \\
            \bottomrule
        \end{tabular}
    \vspace{-0.4cm}
\end{table}

\paragraph{Frame Interpolation}
In subsequent experiments, we utilized the continuous representation of the video data provided by our model to examine the potential for frame interpolation at a desired upsampling rate. To generate additional frames, the \ourdens{} were sliced at uniform intervals between each pair of consecutive frames. Figure~\ref{fig:interp} compares the results obtained by \our{} and VGR~\cite{sun2024splatter} on a selected video object from the DAVIS dataset. The qualitative study reveals that interpolations using our method yield superior results. However, it should be noted that the source code for VGR has not been publicly released by the authors of \cite{sun2024splatter}, preventing a direct comparison of the respective rendering metrics scores.

\paragraph{Video Edition}
To illustrate the adaptability of the \our{} model in editing video data, a series of experiments were carried out on entire scenes and specific objects from the DAVIS dataset. The results, presented in Figures \ref{fig:sim_1} and \ref{fig:sim_2}, confirm that our method allows for both global modification (e.g., multiplication or scaling) of selected objects and for the choice of a single frame to modify some of its elements.

\paragraph{Ablation Study}
In our ablation study, we examined the impact of different hyperparameters of the \our{} model trained on the Bunny dataset \cite{bunny}. Table~\ref{table:bs-pol} presents the final rendering quality metric scores and the number of Gaussians obtained for various batch sizes and degrees of the polynomial function $f$. In turn, Table~\ref{table:init-gauss} presents the final metric values, numbers of Gaussians, and training times with reference to the initial numbers of Gaussians. As can be observed, \our{} attains superior results when applied with a batch size of 3 and a polynomial degree of 7, with a starting number of 0.50M Gaussian components.

\section{Conclusions}

In this paper, we propose the \our{} model, which has been designed for the processing of video. To construct \our{}, we have introduced a novel family of Folded-Gaussian distributions, which allow for the capture of nonlinear patterns in the video stream. The results of the conducted experiments demonstrate that our method enables superior reconstructions and realistic modifications within video frames.

{
    \small
    \bibliographystyle{ieeenat_fullname}
    \bibliography{main}
}

\end{document}